\documentclass[letterpaper]{article}

\usepackage{natbib,alifeconf}  
\usepackage{amsmath}
\usepackage{amssymb}
\usepackage{url,hyperref,cleveref}
\usepackage{booktabs}

\hypersetup{hidelinks}

\usepackage{caption}
\usepackage{subcaption}

\title{Modular Growth of Hierarchical Networks: Efficient, General, and Robust Curriculum Learning}

\author{
    Mani Hamidi$^{1, 2,*}$,
    Sina Khajehabdollahi$^{1}$, 
    Emmanouil Giannakakis$^{1,2}$,
    Tim J. Schäfer $^{1,2}$, \\
    \Large{Anna Levina$^{1,2}$, 
    Charley M. Wu$^{1,2}$}
    \mbox{}\\
    $^1$Department of Computer Science, University of T\"ubingen, T\"ubingen, Germany \\
    $^2$Max Planck Institute for Biological Cybernetics, T\"ubingen, Germany \\
   $^*$mani.hamidi@uni-tuebingen.de 
} 

\begin{document}

\maketitle

\begin{abstract} 

Structural modularity is a pervasive feature of biological neural networks, which have been linked to several functional and computational advantages. Yet, the use of modular architectures in artificial neural networks has been relatively limited despite early successes.  Here, we explore the performance and functional dynamics of a modular network trained on a memory task via an iterative growth curriculum. We find that for a given classical, non-modular recurrent neural network (RNN), an equivalent modular network will perform better across multiple metrics, including training time, generalizability, and robustness to some perturbations. We further examine how different aspects of a modular network's connectivity contribute to its computational capability. We then demonstrate that the inductive bias introduced by the modular topology is strong enough for the network to perform well even when the connectivity within modules is fixed and only the connections between modules are trained. 
Our findings suggest that gradual modular growth of RNNs could provide advantages for learning increasingly complex tasks on evolutionary timescales, and help build more scalable and compressible artificial networks.
\end{abstract}

\section{Introduction}

From bacteria \citep{andrews1998bacteria}, to brains \citep{sporns2016modular}, to man-made artifacts \citep{lake2015human}, many things are composed of modular components, specialized for different purposes and capable of being recombined in distinct configurations to solve new problems.  
In cognitive science, modularity \citep{fodor1983modularity} has played an important role in understanding intelligent behavior as composition of modular, symbolic representations \citep{rubino_hamidi_dayan_wu_2023, zhou2024harmonizing} while
evolutionary accounts have explored how selection pressure towards reducing connection costs may favor modular solutions \citep{clune2013evolutionary}. 
By constraining the search space \citep{happel1994design} and favoring more computationally efficient solutions \citep{yuan2024accelerated}, modular architectures \citep{amer2019review} offer a complementary and more biologically plausible account of intelligence \citep{cosmides1997modular}, at a time when current trends in deep learning have pursued scale at all costs \citep{shen2023efficient}.

Taking inspiration from the modular duplication of entire body parts during gene duplication events \citep{Garcia-Fernandez2005-lj}, we explore the functional utility of modular growth in adapting a recurrent network to a memory task of gradually increasing complexity. We find that coordination of modular growth together with a learning curriculum facilitates a surprising array of advantages, both in terms of performance and costs. 

Specifically, we focus on the domain of multi-timescale signal processing~\citep{Bathellier2008,panzeri_sensory_2010,safavi_signatures_2023}. In tasks such as speech recognition~\citep{graves2013speech}, time-series prediction~\citep{chung2014empirical, torres2021deep}, and navigation \citep{Ari2016}, both biological and artificial agents commonly need to represent and remember relatively long timescales in the underlying network dynamics. 
In artificial neural networks, such dynamics can arise not only via the training of recurrent connectivity, but also by explicitly training the timescales of individual neurons~\citep{perez-nieves_neural_2021,tallec_can_2018, quax_adaptive_2020,yin_effective_2020, fang_incorporating_2021,smith_simplified_2023}. Trainable timescales have been linked with improved performance for rate-~\citep{tallec_can_2018, quax_adaptive_2020} and spiking networks~\citep{yin_effective_2020, fang_incorporating_2021,perez-nieves_neural_2021}.
However, recent work has suggested that greater reliance on connectivity, rather than trainable timescales, is associated with superior performance and robustness of RNNs in similar memory tasks \citep{khajehabdollahi2024emergent}. 
Thus, these current debates suggest that the interaction between trainable and connectivity-based mechanisms, together with their dependence on architectural and training decisions remains relatively poorly understood, even in simple networks performing rather trivial tasks.

One such architectural distinction between artificial and biological networks is the presence of modular topologies \citep{litwin-kumar_slow_2012,chaudhuri_diversity_2014,zeraati_intrinsic_2022,shi_spatial_2022}, which organize various neuronal types into higher level clusters with different timescales \citep{greengard_neurobiology_2001} and specific local connectivity patterns \citep{schaub_emergence_2015}. 
In particular, hierarchical structures in the cortex are associated with a gradual increase in intrinsic timescales \citep{Murray2014, Manea2022} that has been linked to the computational requirements of the tasks performed by different cortical regions \citep{Murray2014}. 

Although artificial hierarchical networks have been used for temporal tasks in the past \citep{Hihi1995hierarchical}, they only explored shallow and static hierarchies that lacked curriculum-based growing mechanisms. Thus, to our knowledge, the role of hierarchical structure in accommodating the emergence of longer timescales has yet to be investigated.

\begin{figure}[t]
    \centering
    \includegraphics[width=0.95\linewidth]{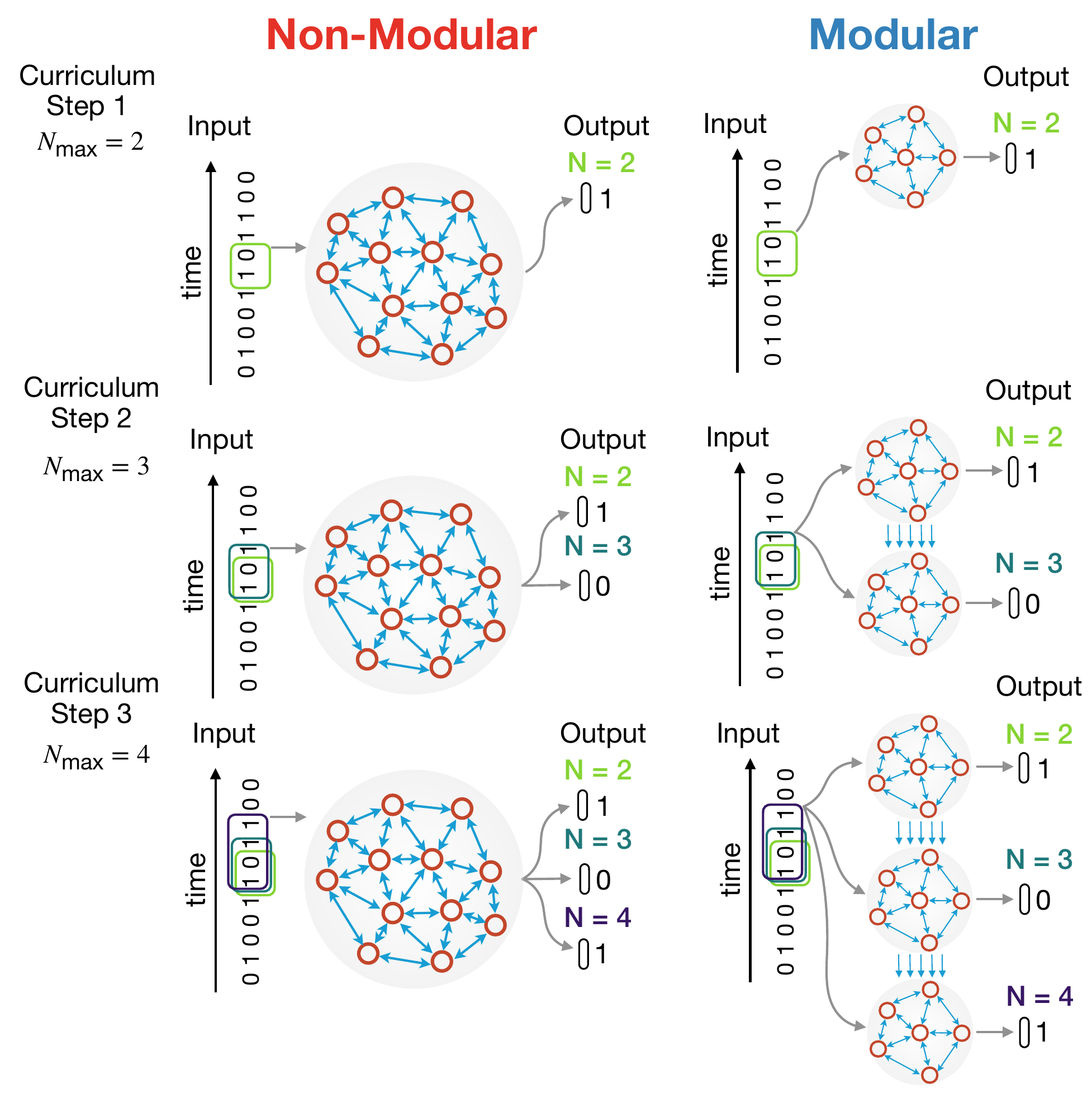}
    \caption{Network Structures. \textit{Non-modular} networks consist of a fixed number of neurons ($M = 20, 54, 91, 128$), with connections that are retrained at each curriculum step. After a given accuracy is reached for a task of length $N$, a new readout head (consisting of 2 neurons) for $N + 1$ is added and the network is retrained for all previous $N$. In contrast, the \textit{modular} network adds a much smaller RNN module ($M_m = 5, 10, 15, 20$) for every readout head that is added in the curriculum. Thus, each readout head is attached to a separate RNN module, rather than one large reservoir as in non-modular networks.}
    \label{fig1}
    \vspace{-1.5em}
\end{figure}

Here, we address this gap by comparing standard RNNs (non-modular) to growing modular networks, using memory tasks with increasing complexity (Fig.~\ref{fig1}).
Building on the success of different curriculum-based training strategies that support the learning of longer timescales \citep{khajehabdollahi2024emergent}, we additionally introduce module-level duplicative growth of the network at every step along the curriculum. This modular strategy is loosely reminiscent of growth over evolutionary timescales that gradually accommodate adaptation to more complex tasks and environments \citep{Lui2011}. Our networks achieve superior performance and robustness, in line with other studies that use neural growth models to train RNNs in a different domain \citep{Najarro2023selfassembly}. 

Our findings suggest that the well-calibrated growth of structured networks can significantly reduce the number of trainable parameters and training steps required to solve complex temporal tasks.

\section{Methods}

We train non-modular and modular networks on the $N$-parity task at increasing complexity levels $N$ and compare their performance, robustness, and learning dynamics. Models are trained using back-propagation-through-time with a stochastic gradient descent optimizer, a cross-entropy loss function, and curriculum learning.

\subsection{Task}
An $N$-parity task is a memory task that is commonly used to assess the capabilities of recurrent network architectures \citep{stork1992solve, hohil1999solving}. The task requires the accurate retention of a sequence of binary digits to perform a modulo 2 summation over the last $N$ digits in the sequence. Therefore $N$ captures the complexity of the task. 
The input is a random binary sequence $S$ with length $L$ chosen uniformly from the interval $\{N + 2, 4N\}$, with one bit provided to the network at each time step. The network must output the binary sum ($\mathrm{XOR}$) of the last $N$ digits. To update the output at every time step, the network must learn to store some representation of the values and order of the last $N$ digits in memory. 

Although simple, this task provides a foundation for testing the representation learning and memory capacities of artificial neural networks, while simultaneously allowing us to control the difficulty of the task 1 bit at a time using $N$. Furthermore, multiple $N$-parity tasks can be computed for a single sequence, allowing the possibility of training concurrent tasks/readout layers on the same inputs. This concurrent training is used to encourage more universal features that can be shared between the different tasks and to prevent catastrophic forgetting. 

The success criterion for completing task $N$ is therefore defined as jointly satisfying i) $>98\%$ accuracy on task $N$ and ii) maintaining an average of $>98\%$ accuracy on all previous tasks.
Accuracy is reported as an average over multiple tests, with random chance yielding $50\%$. We report network performance using $N_\text{solved}$ which describes the largest task that it was able to solve after $60$ epochs.

\begin{figure*}[t]
    \centering
    \includegraphics[width=0.9\textwidth]{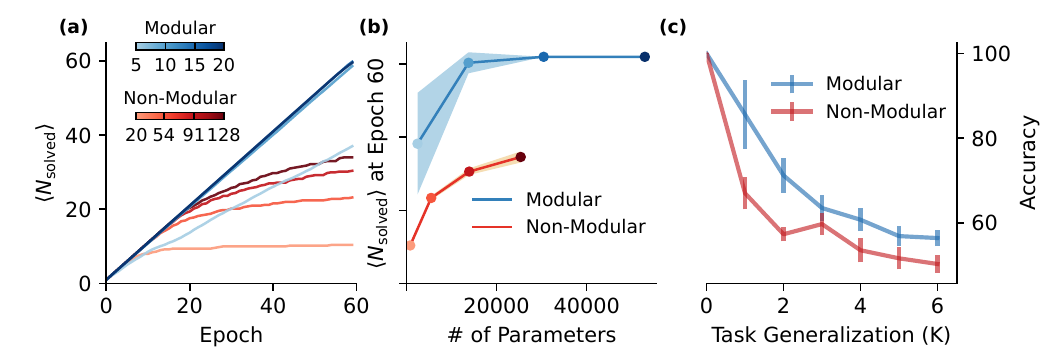}
    \caption{Performance comparison of modular and non-modular architectures with different numbers of neurons (color bars) allocated to their recurrent processing unit. For modular networks, the number of neurons is reported per module. For non-modular networks, the number indicates the total number of neurons in the single recurrent core. These numbers were chosen to allow for a comparable total number of learnable parameters for a given task difficulty $N$. (\textbf{a}) Learning curves show that modular architectures with sufficient ($> 5$) neurons can solve a new task at every epoch while the non-modular networks plateau in their learning ability. (\textbf{b}) Pareto frontier of performance after 60 training epochs, shows that the modular architecture always achieves better performance for the same number of parameters. Note that the saturation of performance at $N_{\text{solved}}=60$ for the modular architecture is due to the choice of training epochs.
    (\textbf{c}) Generalization performance of networks trained on a task difficulty of $N$ and then tested on tasks of $N+K$. Accuracy is measured as the percentage of correct trials and error bars indicate the standard deviation. Results are averages over 4 non-modular and 3 modular networks, and errorbars show STD.}
    \label{fig2}
    \vspace{-1em}
\end{figure*}

\subsection{Networks} 
We use simple rate neurons with a trainable time constant $\tau$ whose activity is defined by: 
\begin{equation}
r_i(t) =   \left(1 - \frac{\Delta t}{\tau_i}\right) \cdot r_i(t-\Delta t) + \frac{\Delta t}{\tau_i} \cdot \left[C_t \right]_{\alpha}
\label{eq:rnn}
\end{equation}
$C_t$ is the input at each timestep $t$ and the non-linearity $[\cdot]_{\alpha}$ is the leaky ReLU function with negative slope $\alpha$, given by:
\begin{equation}
[x]_{\alpha} = \left\{
        \begin{array}{ll}
            x, & \ x \geq 0 \\
            \alpha \cdot x, & \ x < 0.
        \end{array}
    \right.
\end{equation}
During learning, we train both the network connectivity (recurrent and feedforward) as well as the individual neuron timescales $\tau_i$. The trainable timescale $\tau_i$ has a minimum value of 1 indicating a neuron that reacts only to the shortest timescales (i.e., the current input $C_t$), whereas larger values integrate information from longer timescales in the past.  

\subsubsection{Non-modular networks.}
In non-modular networks \citep{jaeger2002tutorial}, the input at each timestep $t$ for  neuron $i$ is: 
\begin{equation}
C_t = \sum_{ j \neq i}^{M} W^R_{ij} \cdot r_j(t-\Delta t) + W^I_{i} \cdot S(t) + b_i.
\end{equation}
Here, $W^R$ is the recurrent connectivity, $W^I$ is the feedforward input connections, $S(t)$ is the input signal, $b_i$ is the neuron bias, and time discretization $\Delta t=1$.

The non-modular networks follow the multi-head curriculum in \citep{khajehabdollahi2024emergent}: At the first curriculum step, a single linear readout head is trained to solve the task for $N = 2$. Upon successfully completing task $N$, a new readout head is added and is trained to solve the $N+1$-parity task. Previous readout heads continue to be trained after every step in the curriculum. Thus, at the $m_{th}$ step of the curriculum, the network has $m$ readout heads solving the task for $N = 2, \dots m - 1$.

\subsubsection{Modular networks.}

In modular networks, each neuron receives input from the other neurons of the same module, and for modules $m > 1$, neurons also receive input from a feed-forward connection from the previous module $m-1$:
\begin{equation}
\begin{aligned}
C_t^m ={} & \sum_{ j \neq i}^{M_m} W^R_{ij} \cdot r_j^m(t-\Delta t) \\
+ & \sum_k^{M_m} W^{FF}_{ik} \cdot r_k^{m-1}(t-\Delta t) \\
+ & \; W^I_{i} \cdot S(t) + b_i
\end{aligned}
\end{equation}
where $W^{FF}$ is the feed-forward connectivity from the previous module and $r_k^{m-1}(t - \Delta t)$ is the activity of neuron $k$ in module $m - 1$ from the previous timestep. $M_m$ indicates the number of neurons in each module, which is fixed to be one of $M_m \in [5, 10, 15, 20]$ in our experiments. 

For training the modular networks, we follow a growing curriculum as follows: Starting with a single module with a small population (5, 10, 15, or 20 neurons), we train a linear readout for $N = 2$. Then, at each curriculum step, we add a new module and readout head for each new $N$. The connectivity of the new module is a copy of the corresponding (feedforward and recurrent) connectivity for the $N - 1$ module and receives feedforward input from it. Thus each module is specialized to solving the task for a single $N$. At each curriculum step, we freeze all network connections for all modules except the last and train only the recurrent and feedforward connections (input from sequence and input from previous module) of the last module.

\subsection{Timescale Estimation}

Apart from the trainable parameters $\tau_i$ that define the timescale of each neuron explicitly, we also compute the \textit{effective} timescale of a neuron, which is influenced by its intrinsic trained $\tau_i$ and the extrinsic modulation of its activity through its connections with the rest of the network. We determine this network-mediated effective timescale for individual neurons by calculating the lagged autocorrelations (AC) of their activity during task performance, following previously developed methods \citep{khajehabdollahi2024emergent}. The AC of each neuron is then modeled with an exponential function featuring one or two timescales, selecting the best-fit model based on the Akaike Information Criterion \citep[]{Akaike1974}. To avoid estimation bias, we utilize long time series ($10^5$ time steps) of activity \citep{zeraati_flexible_2022}.
\subsection{Robustness}

We evaluate robustness by measuring the accuracy of the network after perturbing one of its three trained parameter groups, $W^R$, $W^{FF}$ or $\tau$.
We define the magnitude of the perturbation $\varepsilon_W$ as a function of the magnitude of the weights~\citep{wu_adversarial_2020}:

\begin{equation}\label{eq:perturb}
    \widetilde{W} = W + \varepsilon_W \frac{\xi_W}{||\xi_W||}||W||
\end{equation}
where $\xi_W \sim \mathcal{N}(0, \mathbb{I}^{n \times n})$ and $||\cdot||$ represents the Frobenius norm. This normalization allows for comparable amounts of perturbation across networks of different types and sizes. We also only perturb
$\tau_i$ in positive direction to avoid $\tau < 1$.   

\section{Results}

Our findings are divided into two main sections. In the first part, we demonstrate that modular networks following an iterative growing curriculum outperform equivalent non-modular RNNs in terms of task performance, training speed, generalizability, and robustness to perturbation of their learned connectivity. In the second part, using perturbation and ablation techniques, we investigate how different aspects of modular networks (feedforward vs. recurrent connectivity and trainable vs. effective single-neuron timescales) contribute to their computational capabilities.

\subsection{Modular vs. Non-Modular Networks}

\subsubsection{Performance and Generalization} 

We first compare a set of modular networks with different module sizes ($M_m = 5, 10, 15, 20$) against a set of non-modular networks  ($M = 20, 54, 91, 128$) with an equivalent number of trainable parameters (at $N=30$). 
Performance is calculated as the average task difficulty solved, $N_\text{solved}$, both as a function of training time (Fig.~\ref{fig2}a) and the number of parameters (Fig.~\ref{fig2}b).

\begin{figure}[tb]
    \centering

        \includegraphics[width=\linewidth]{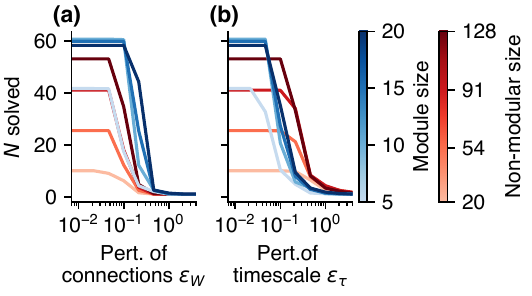}

    \caption{
    Weight perturbations degrade performance in both modular and non-modular networks of different sizes (colours). \textbf{(a)} Modular networks are more robust to perturbation of connections (modular networks: feedforward and recurrent weights).
    \textbf{(b)} Non-modular networks are more robust to the perturbation of single-neuron trained timescales.
    }
    \label{fig:perturb1}
    \vspace{-1em}
\end{figure}

Overall, we find that all modular networks reach a higher task difficulty ($N_{\text{solved}}$) than the corresponding non-modular networks (Fig.~\ref{fig2}a). Except for the smallest modular networks ($M_m=5$), all modular networks maintain a steady, linear progression through the curriculum, solving every task after a single epoch of training. As an extreme test of the limits of the modular network, we were able to reach a maximum of $N=200$ in one network before ending the simulation, with even greater capabilities being theoretically possible given sufficient training time. 
The non-modular networks, on the other hand, plateau much earlier in their progression through the curriculum. Thus, for a fixed number of trainable parameters, our results suggest there is a modular architecture that is capable of solving more complex tasks with the same amount of training.

We also explore the impact of a limited budget for learnable parameters as an alternative evaluation of the cost-benefit trade-off between two architectures. Figure~\ref{fig2}b shows the Pareto frontiers of the architectures after 60 epochs, indicating that a modular architecture can solve a task with fewer parameters than a non-modular network.

Finally, we test generalization performance on networks of size $M_m = 15$ and $M=91$, by training them to $N_{solved} = 10$ and then test their ability to solve $N = N_{solved} + K$, for $K = 1, 2, \dots$. For modular networks, the $N=10$ module is now tested for solving $N=10 + K$, while the non-modular network is tested on a new read-out head for $N=10 + K$. We only allow 10 epochs of additional training on the new task. The results are shown in Figure~\ref{fig2}c, where although performance decays with $K$ for both types of networks, modular networks consistently reach a higher accuracy compared to non-modular networks, thus demonstrating superior generalization.

\subsubsection{Robustness to Perturbations.}

Next, we tested the robustness of both network architectures to perturbations of connectivity weights (Fig.~\ref{fig:perturb1}a) and trained timescale parameters (Fig.~\ref{fig:perturb1}b). The perturbations were proportional to the magnitude of the weights of the layer being targeted (Eq.~\ref{eq:perturb}). For different perturbation levels, we measure the network performance via the number of tasks that the network is still able to solve with accuracy $> 0.9$. 

Overall, Figure~\ref{fig:perturb1} shows the degradation of performance as a function of the size of the perturbation on the connections (both recurrent $W^{R}_{ij}$ and feedforward $W^{FF}_{ij}$ in the case of the modular networks) and the timescale parameters ($\tau_i$) of the network. Modular networks not only have better initial performance but are also more robust against connectivity perturbations (Fig.~\ref{fig:perturb1}a). This is visible in the later inflection point, reflecting how larger perturbations are required to degrade the performance of modular networks.
However, modular networks are more sensitive to the perturbation of time-scale parameters (Fig.~\ref{fig:perturb1}b). 
A special case is modular networks with $M_m=5$ that appear to be more sensitive to both perturbations than their larger counterparts. 
For all other cases, the number of neurons does not significantly impact the robustness of either network type. So the inflection point of degrading performance is an inherent feature of network architecture. In sum, modular networks are significantly more robust against connectivity perturbations, but are more sensitive to perturbation of time-scale parameters.

\subsection{Functional Analysis of Modular Networks} 

\subsubsection{Trained vs. Effective Timescales.} 

\begin{figure}[tb]
    \centering

        \includegraphics[width=\linewidth]{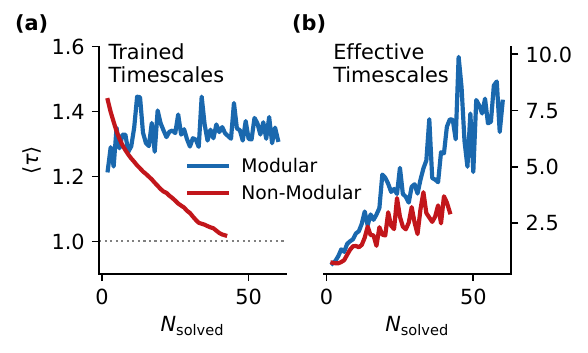}
        \caption{Change in trained and effective timescales for different $N_{solved}$. (\textbf{a}) The trained timescale of the modular network stays the same across modules, while the timescales of the non-modular network converge to 1 as the $N_{solved}$ increases. (\textbf{b}) The effective timescales of both networks increase steadily with $N_{solved}$. A modular network with a module size of 15 neurons and a non-modular network with an equal number of trainable parameters were used. Results are averages over 4 networks.}\label{fig:timescales}
        \vspace{-1em}
\end{figure}

To better understand the increased sensitivity of modular networks to timescale perturbations, we examine the mechanism by which long timescales emerge in the networks. We do so by examining both the \textit{trained timescale} parameterized by $\tau_i^m$ (Fig.~\ref{fig:timescales}a) and also the \textit{effective timescale} inferred from the activity of each neuron  (Fig.~\ref{fig:timescales}b). 
In the modular case, the average trained and effective timescales of neurons are reported separately per module solving $N=N_\text{solved}$. In the non-modular case, the average is taken over all neurons in the entire network at the time when it has just solved the $N=N_\text{solved}$ task.

First, we find that neurons in modular networks maintain stable trained timescales, $\langle\tau_i \rangle_m$, throughout the curriculum, but non-modular networks decay to the fastest possible rate of one (Fig.~\ref{fig:timescales}a).
In both cases, these trained timescale values seem too small to account for the long timescales that are required for the large $N$ tasks they can solve. This motivated us to examine the effective timescales, which are a product of the connectivity structure of the whole network. 

Figure~\ref{fig:timescales}b shows the increase in the average effective timescales of neurons in both modular and non-modular networks for increasingly complex tasks $N$. This observation is consistent with recent observations \citep{khajehabdollahi2024emergent} suggesting a circuit-level implementation of memory by learning appropriate connection weights ($W$) rather than learning slow timescales ($\tau_i$) at the single neuron level. Notably, neurons in the modular networks appear to harbor memory of longer timescales compared to the non-modular counterparts at the same stage in the curriculum.

\begin{figure*}[htb]
    \centering
    \includegraphics[width=0.9\textwidth]{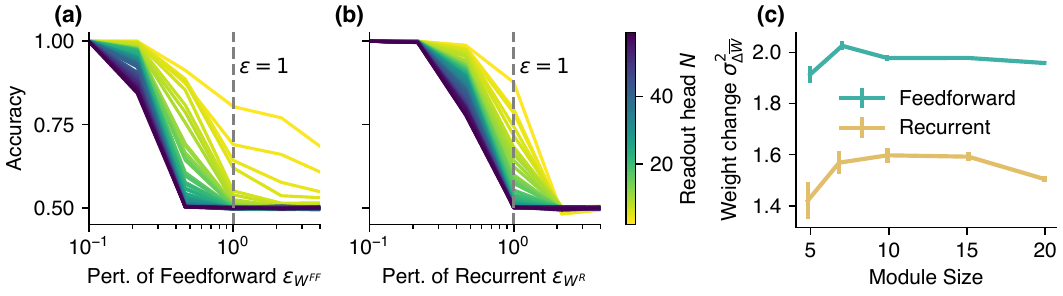}
    \caption{
    Feedforward connections are more sensitive than recurrent connections in modular networks. Here, we use $M_m=15$ but achieve qualitatively similar results for other network sizes. 
    \textbf{(a)} Perturbing feedforward weights affect downstream modules more strongly than earlier modules.
    \textbf{(b)} The recurrent weights exhibit a similar qualitative pattern but are quantitatively more robust against the same levels of perturbations.
    \textbf{(c)} Variability of feedforward versus recurrent connection weights across modules, where variability is inversely related to the degree of conservation (i.e., the amount of shared weights from one module to the next).
    Accuracy is averaged over 5 networks, 10 perturbations, and 1000 continuous evaluations. Error bars show SEM.
    }
    \label{fig:perturb2}
    \vspace{-1em}
\end{figure*}

\subsubsection{Recurrent vs Feedforward Connections.}

We now turn to how the topological constraints imposed by the modular architecture suggest different qualitative roles for feedforward and recurrent connections. 
For instance, the feedforward connections serve as the only bottleneck through which the recurrent computations of each module are reused by subsequent modules (Fig.~\ref{fig1}), making them potentially more vulnerable to noise, whereas the recurrent connections may be more robust.
Furthermore, the dependence of long effective neural timescales on appropriate connectivity patterns, encourages us to examine the distinct roles that feedforward ($W^{FF}$) and recurrent ($W^R$) weights play in performance. 

To do so, we performed separate perturbation analyses targeting either feedforward $W^{FF}$ or recurrent connections $W^R$. We evaluate performance by measuring the accuracy of each module's prediction for its corresponding task $N$.

Figure~\ref{fig:perturb2}a-b shows that the feedforward connections are more sensitive to perturbations relative to the recurrent weights.
Furthermore, modules corresponding to more difficult tasks (indicated by darker colors; Fig.~\ref{fig:perturb2}a-b), suffer more because they are affected by the cumulative effect of all the perturbations to preceding modules.
Note that the effect of a perturbation is always downstream, due to the sequentially connected architecture of modules, such that a single perturbation equally affects all downstream modules.

To understand the basis for the sensitivity of $W^{FF}$, we hypothesize that $W^{FF}$ are subject to more fine-tuning via back-propagation, in order to facilitate specialization to each new task. On the other hand, we expect $W^{R}$ to be more conserved, and subject to less modification from one task to the next.
We define the change in the connection weight between nodes $i$ and $j$ across two consecutive modules as $\Delta \overline{W_{ij}}$, where we normalized each weight by its mean and standard deviation to correct for systematic differences across module sizes \citep{LeCun2012-xo}.
The variance of the normalized weight change, $\sigma^2_{\Delta \overline{W}_{ij}}$, is the change in the normalized weight, $\Delta \overline{W_{ij}} = \overline{W}_{ij}^{m} - \overline{W}_{ij}^{m - 1}$ between corresponding neurons, $i, j$, in consecutive modules, $m, m-1$. This serves as an inverse proxy for the degree to which the weights are conserved during training, where lower variance corresponds to greater conservation of weights.  

Figure~\ref{fig:perturb2}c shows how recurrent weights are more conserved than feedforward weights (lower variance). This supports the hypothesis that sensitivity of feedforward connections to perturbations is due to a more general sensitivity to modification, both from noise and from back-propagated error during training.

\subsubsection{Weight Freezing.}

The low variability of the recurrent weights motivated us to test the degree to which a network can continue to learn despite a complete freezing of either its recurrent or the feedforward weights, after passing the very first step in the curriculum.
A freezing of weights corresponds to a reduction of computational costs, by converting learnable parameters of a model into fixed biases. In evolutionary biology, this is known as the Baldwin effect \citep{gruau1993adding}, where plastic behaviors become fixated in response to environmental stability.
In the case of frozen recurrent weights, we trained the first module for $N = 2$ and then duplicated the learned recurrent weights on all subsequent modules, training only the feedforward connections between modules. Conversely, in the case of frozen feedforward weights, we train the feedforward connections between the $N = 2$ and $N = 3$ modules and then duplicate this trained connectivity between all subsequent modules, only training the internal recurrent connections of each module.

\begin{figure*}[htb]
    \centering
    \begin{subfigure}[b]{0.43\textwidth}
        \includegraphics[width=\textwidth]{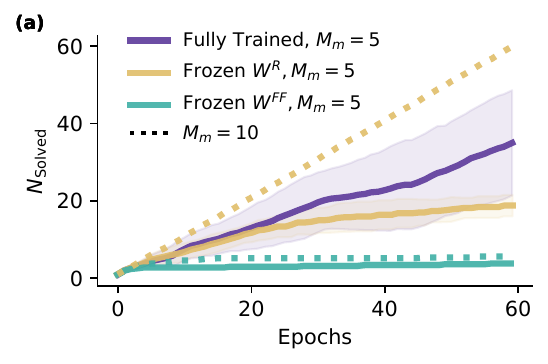}
    \end{subfigure}
    \begin{subfigure}[b]{0.43\textwidth}
    \includegraphics[width=\textwidth]{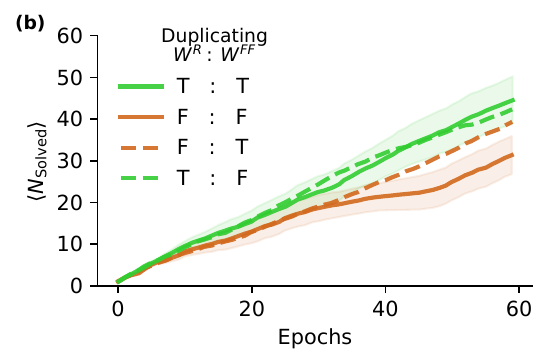}
    \end{subfigure}
\caption{The impact of weight freezing and duplication on feedforward and recurrent weights. (\textbf{a}) Weight freezing. Solid lines are a modular network with $M_m=5$, selected for the strongest differences. Here, we can see that freezing recurrent connections (orange)  after solving the first task ($N=2$) still needs to relatively good performance, whereas freezing feedforward connections (teal) severely impairs learning. In networks of larger module size ($M_m=10$; dotted lines), the difference between frozen recurrent weights and a fully trained network disappears (both dotted orange), while frozen feedforward weights remain impaired. (\textbf{b}) Weight duplication. New modules are initialized with either a duplicated copy (T) of their weights as they appeared in the previous module, or a random initialization of weights (F). Duplication of recurrent weights confers a slight advantage (green), while either duplicative or randomly initialized feedforward weights both confer a slight impairment to performance. Results are averages over 5 networks and shaded regions indicate SEM. }
    \label{fig:reservoir}
    \vspace{-1em}
\end{figure*}

Figure~\ref{fig:reservoir}a shows that networks with frozen feedforward weights fail catastrophically to solve the task for any $N \geq 3$. However, equivalent networks with frozen recurrent connectivity \citep[resembling deep reservoirs;][]{Gallicchio2021}, perform much better, comfortably reaching an $N_{solved} \approx 15$ before stagnating (Fig.~\ref{fig:reservoir}a) and even outperforming non-modular networks that taper at $N \approx 10$. 
Thus, while both network types underperform the original network for which both recurrent and feedforward connections are trained, we see that, in line with our previous results, the learnability of the feedforward connections appears to be much more important for task performance. 

\subsubsection{Weight Duplication.} 

The weight freezing experiments involved the reuse (or duplication) of weights from the previous module followed by a complete blockage of their training. Here we consider duplicative versus random initialization of the weights, and maintain the plasticity of weights during training. 
Theoretically, duplication can be beneficial by amortizing the costs of learning through the reuse of already-trained weights, or it could be detrimental by initializing the network in a state unsuitable for the new task. 
Empirically, previous work using weight-agnostic neural networks \citep{gaier2019weight}, compositional pattern-producing networks \citep{stanley2007compositional}, and hypernetworks \citep{ha2016hypernetworks}, have demonstrated the utility of similar ``weight-sharing'' schemes that support high-performing neural networks with a fraction of the parameter count. Motivated by these past successes, here we explore whether the duplication of recurrent or feedforward components can accelerate the training of the network, or if such methods are detrimental by trapping the behavior of the network in a local optima.

Figure \ref{fig:reservoir}b shows the results of these experiments, where weights were either duplicated (T for true) or not (F for false); in case of the latter, they were initialized randomly following a uniform distribution. The lines show the average performance of 6 networks in each of the 4 different conditions, with and without duplication of feedforward $W^{FF}$ and/or recurrent $W^{R}$ weights. These results show that the duplication of recurrent weights $W^R$ appears to account for an unambiguous advantage (Fig.~\ref{fig:reservoir}b; green curves), compared to the duplication of feedforward weights, which do not confer any additional benefits. Indeed, duplicating feedforward weights confers little to no improvement to when neither weights are duplicated (Fig.~\ref{fig:reservoir}b, brown curves).  
We also experimented with noisy duplication, but given the weak overall effect, we only show the results of exact duplication, without noise. Furthermore, different duplication strategies mattered less at larger module sizes, so we focused only on networks with five nodes per module $M_m=5$.

These results are consistent with our previous analysis showing that feedforward connections must diverge from their past configurations to specialize for their respective task. Therefore, as long as the feedforward connections remain plastic after a growth event, they are molded strongly by the error signal from their new task, making their initial state irrelevant. In contrast, we have already shown how keeping the same or similar recurrent weights across multiple tasks is relatively adaptive. Thus, duplication simply initializes the weights near their last functional state, eliminating the need to retrain them to that stage at every epoch, and allows for minimal fine-tuning to improve their adaptiveness over longer timescales.

\section{Discussion}

In sum, we find that modular networks, trained via a growing curriculum to solve a memory task of adjustable difficulty, outperform standard recurrent neural networks (RNNs) of fixed size in terms of accuracy, training time, generalization, and robustness to perturbations of the learned connectivity.

Biological neural networks are highly structured \citep{hagmann2008mapping, sporns2004small, sporns2016modular}, with multiple neuron types each following distinct connectivity patterns \citep{peng2021morphological, liu2023neuronal} that form complex circuits repeating across brain regions \citep{douglas2004neuronal, shepherd2011microcircuit}. The repetition of similar complex structures in the brain potentially enables the development of very complex neural circuits from limited genetic information \citep{rakic2009evolution, geschwind2013cortical, stanley2007compositional}, and may also play a role in more efficient information processing \citep{bassett2006small, Sporns2013}. Moreover, inhomogeneous network topology has been shown to generate non-trivial dynamics \citep{litwin-kumar_slow_2012} and act as an inductive bias for local learning \citep{Myself2023}. Here, our findings add to this literature by suggesting relatively simple topological modularity can be utilized to align network dynamics with task requirements, thus increasing performance and reducing training time and cost.

Our modular architecture allows us to study the roles played by different aspects of the network’s structure. In particular, we focus on the distinction between the recurrent weights within each module and the feedforward weights connecting adjacent modules. Our analysis indicates that the training and task specialization of feedforward weights between modules are vastly more important than the recurrent connections within each module. This suggests that the function of the recurrent connectivity is largely limited to creating appropriate and generic dynamical units (and can thus be effectively reused through duplication), while the feedforward connections control the flow of information that is necessary for solving the specifics of a given task.

While our network included trainable single-neuron timescales, which have been utilized in RNNs solving temporal tasks \citep{tallec_can_2018, quax_adaptive_2020, khajehabdollahi2024emergent}, we find that their importance is minimal in terms of network performance. The distribution of trained timescales barely changes across modules, suggesting that the network fully relies on its connectivity (particularly the hierarchical topology) for developing the long timescales necessary for solving temporal tasks. This augments previous findings  suggesting that network topology can generate long timescales with relatively random connections \citep{khajehabdollahi2024emergent}.

This understanding of the relative importance of different network components, allows us to test whether training even fewer parameters can lead to comparable results. We find that training the networks as deep reservoirs \citep{Gallicchio2021} leads to reasonably good performance despite reducing the trainable parameters by half. This suggests that functional sub-networks can be used as building blocks in larger systems with little or no training, which seems like a promising avenue for future research. 

The importance of initializing neural networks in an appropriate dynamical regime has been widely explored \citep{Zierenberg2020, Khajehabdollahi2022}, and our findings here suggest that hierarchical modular structures could be used to generate networks with beneficial dynamics for learning temporal tasks. Recycling functional structures and incorporating them in complex networks of interacting sub-units is a widespread characteristic of biological networks \citep{felleman1991distributed} and has shown promise in artificial settings \citep{happel1994design, sharkey1996combining, amer2019review}. Our findings suggest that this approach is not only more efficient in terms of the number of trainable parameters, but it can also boost performance as well as the ability to generalize.

\subsubsection{Limitations \& Future Work.}

A basic premise of our work was that the power and cost of a neural network are proportional to the number of its trainable parameters. The number of parameters, therefore, provided a basis for comparing neural networks of two different architectures. Our results indicate that one can maintain or even exceed performance while being more frugal in resource expenditure on connections. While this result is relevant both for artificial \citep{amer2019review} and biological networks \citep{clune2013evolutionary}, where these connections carry non-trivial costs, it does overlook the relative cost of producing and maintaining neurons in addition to their connections. Our modular architecture incurs a heavy cost in terms of individual neural processing units, scaling linearly with task complexity in proportion to each module's size. Indeed, the ability of modular networks to learn longer effective timescales points to the importance of large but sparse networks in solving a memory task. Future work could address the trade-off between these two resources and their impact on performance.

Finally, even though we found a functional role for duplicative growth in artificial networks, the observed benefits were modest. In evolutionary contexts, relaxed selection \citep{Deacon2010-of} provides a mechanism by which duplication can accommodate enhanced adaptation by fostering synergistic interactions between duplicated modules. In our current setup, each duplicated module is required to solve a new task, subjecting it to heavy selection pressure that prevents it from exploring the entirety of the weight space. Relaxing this constraint by introducing redundancy in modules that are assigned the same task, could amplify the positive effects of duplication in future research.

In conclusion, modular growth offers a promising avenue for training lighter networks with more scalable properties. In the era of large models with increasingly prohibitive costs, there is great interest in developing new tools to compress models with fewer parameters \citep{Cheung2019-mj} and fine-tune them to perform novel tasks \citep{Hu2021-nz}. Our work, like other evolutionary-inspired approaches \citep{Akiba2024-wm}, can offer new directions towards this end.       

\section{Acknowledgements}
 MH, EG, TS and SK thank the International Max Planck Research School for Intelligent Systems (IMPRS-IS) for their support. MH, EG, and CMW are supported by the German Federal Ministry of Education and Research (BMBF): Tübingen AI Center, FKZ: 01IS18039A. MH and CMW are funded by the Deutsche Forschungsgemeinschaft (DFG, German Research Foundation) under Germany’s Excellence Strategy–EXC2064/1–390727645.
TS is supported by Else Kröner Medical Scientist Kolleg ``ClinbrAIn: Artificial Intelligence for Clinical Brain Research''.

\footnotesize
\bibliographystyle{apalike}
\bibliography{bibliography}

\end{document}